\documentclass{article}

\usepackage{microtype}
\usepackage{graphicx}
\usepackage{subcaption}
\usepackage{booktabs}
\usepackage{hyperref}

\usepackage[preprint]{icml2026}

\usepackage{amsmath}
\usepackage{amssymb}
\usepackage{mathtools}
\usepackage{amsthm}
\usepackage{tikz}
\usetikzlibrary{arrows.meta,positioning,fit,backgrounds,calc}

\usepackage[capitalize,noabbrev]{cleveref}

\theoremstyle{plain}
\newtheorem{theorem}{Theorem}[section]

\theoremstyle{definition}
\newtheorem{definition}[theorem]{Definition}

\icmltitlerunning{Experience Compression Spectrum}

\begin{document}

\twocolumn[
  \icmltitle{Experience Compression Spectrum: Unifying Memory, Skills,\\and Rules in LLM Agents}

  \begin{icmlauthorlist}
    \icmlauthor{Xing Zhang}{aws}
    \icmlauthor{Guanghui Wang}{aws}
    \icmlauthor{Yanwei Cui}{aws}
    \icmlauthor{Wei Qiu}{hsbc}
    \icmlauthor{Ziyuan Li}{hsbc}
    \icmlauthor{Bing Zhu}{hsbc}
    \icmlauthor{Peiyang He}{aws}
  \end{icmlauthorlist}

  \icmlaffiliation{aws}{AWS Generative AI Innovation Center}
  \icmlaffiliation{hsbc}{HSBC Holdings Plc., HSBC Technology Center, China}

  \icmlcorrespondingauthor{Peiyang He}{peiyan@amazon.com}

  \icmlkeywords{LLM Agents, Experience Compression, Agent Memory, Skill Discovery, Efficient Agentic Systems, Scalable Learning}

  \vskip 0.3in
]

\printAffiliationsAndNotice{}

\begin{abstract}
As LLM agents scale to long-horizon, multi-session deployments, efficiently managing accumulated experience becomes a critical bottleneck. Agent memory systems and agent skill discovery both address this challenge, extracting reusable knowledge from interaction traces, yet a citation analysis of 1{,}136 references across 22 primary papers reveals a cross-community citation rate below 1\%. We propose the \emph{Experience Compression Spectrum}, a unifying framework that positions memory, skills, and rules as points along a single axis of increasing compression (5--20$\times$ for episodic memory, 50--500$\times$ for procedural skills, 1{,}000$\times$+ for declarative rules), directly reducing context consumption, retrieval latency, and compute overhead. Mapping 20+ systems onto this spectrum reveals that every system operates at a fixed, predetermined compression level: none supports adaptive cross-level compression, a gap we term the \emph{missing diagonal}. We further show that specialization alone is insufficient (both communities independently solve shared sub-problems without exchanging solutions), that evaluation methods are tightly coupled to compression levels, that transferability increases with compression at the cost of specificity, and that knowledge lifecycle management remains largely neglected. We articulate open problems and design principles for scalable, full-spectrum agent learning systems.
\end{abstract}

\begin{figure*}[t]
\centering
\begin{tikzpicture}[
  >=Stealth,
  level/.style={
    draw, rounded corners=4pt, minimum width=3.1cm, minimum height=1.8cm,
    font=\small, align=center, thick
  },
  sysbox/.style={font=\footnotesize, align=center, text width=3.0cm},
  arrowlbl/.style={font=\scriptsize, midway, above},
]

\node[level, fill=gray!12] (l0) at (0,0) {\textbf{Level 0}\\Raw Trace\\{\scriptsize 1:1}};
\node[level, fill=blue!12] (l1) at (4.2,0) {\textbf{Level 1}\\Episodic Memory\\{\scriptsize 5--20$\times$}};
\node[level, fill=green!12] (l2) at (8.4,0) {\textbf{Level 2}\\Procedural Skill\\{\scriptsize 50--500$\times$}};
\node[level, fill=orange!12] (l3) at (12.6,0) {\textbf{Level 3}\\Declarative Rule\\{\scriptsize 1000$\times$+}};

\draw[->, thick] (l0) -- (l1) node[arrowlbl] {extract};
\draw[->, thick] (l1) -- (l2) node[arrowlbl] {abstract};
\draw[->, thick] (l2) -- (l3) node[arrowlbl] {generalize};

\node[sysbox, below=0.5cm of l0] (s0) {Conversation logs\\Execution traces};
\node[sysbox, below=0.5cm of l1] (s1) {MemSkill, Mem0\\A-MEM, MemoryOS\\ALMA, MemMA\\Memory-R1, Mem-$\alpha$\\MemPO, SSGM};
\node[sysbox, below=0.5cm of l2] (s2) {Voyager, SkillWeaver\\EvoSkill, Trace2Skill\\AutoSkill, SkillRL\\CASCADE, EvolveR};
\node[sysbox, below=0.5cm of l3] (s3) {\emph{(largely empty)}\\Constitutional AI\\(pre-specified only)};

\begin{scope}[on background layer]
  \fill[red!6, rounded corners=6pt]
    ([xshift=-0.3cm, yshift=0.4cm]l1.north west) rectangle
    ([xshift=0.3cm, yshift=-3.8cm]l3.south east);
  \draw[red!40, dashed, thick, rounded corners=6pt]
    ([xshift=-0.3cm, yshift=0.4cm]l1.north west) rectangle
    ([xshift=0.3cm, yshift=-3.8cm]l3.south east);
\end{scope}
\node[font=\footnotesize\bfseries\itshape, text=red!70, anchor=south east]
  at ([xshift=0.2cm, yshift=-3.7cm]l3.south east) {missing diagonal};

\draw[dashed, thick, purple] ([yshift=-2.4cm]l1.south) -- ([yshift=-2.4cm]l2.south)
  node[midway, below, font=\footnotesize, text=purple, text width=3cm, align=center]
  {ExpeL, AutoAgent\\(cross-level, but fixed)};

\draw[->, gray, thick] ([yshift=0.5cm]l0.north west) -- ([yshift=0.5cm]l3.north east)
  node[pos=0.5, anchor=south west, yshift=2pt, font=\small, text=black] {Generalizability $\longrightarrow$};
\draw[->, gray, thick] ([yshift=1.1cm]l3.north east) -- ([yshift=1.1cm]l0.north west)
  node[pos=0.5, anchor=south east, yshift=2pt, font=\small, text=black] {$\longleftarrow$ Specificity};

\end{tikzpicture}
\caption{\textbf{The Experience Compression Spectrum.} Existing agent learning systems map onto a single axis from raw traces to abstract rules. Memory systems cluster at Level~1, skill systems at Level~2, with Level~3 largely empty. A small number of cross-level systems (dashed) bridge Levels~1--2 but none support adaptive level selection. Compression ratios are approximate.}
\label{fig:spectrum}
\end{figure*}
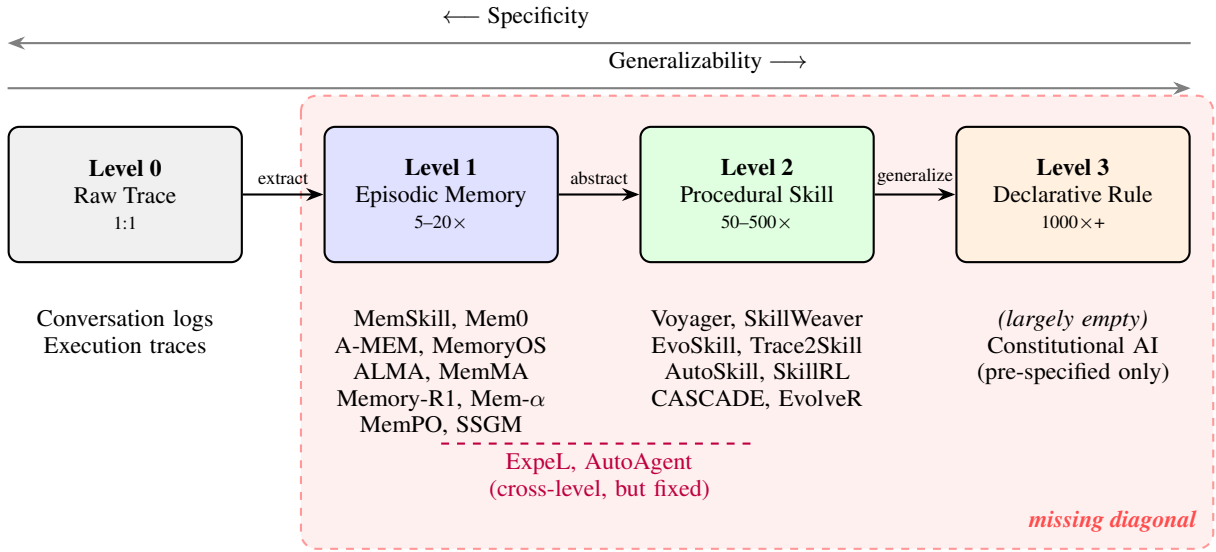

\section{Introduction}
\label{sec:intro}

As LLM agents move from single-session demos to persistent, long-horizon deployments, they accumulate vast interaction experience. An agent handling thousands of tasks per day generates traces that quickly overwhelm any practical context window or retrieval budget, making efficient experience management a first-order scalability challenge. Two research communities have emerged to address this: the \textbf{agent memory} community develops systems for extracting and retrieving experiential knowledge \citep{hu2025memorysurvey,kang2025memoryos,packer2023memgpt}, while the \textbf{agent skill} community builds frameworks for discovering and reusing procedural capabilities from execution traces \citep{wang2023voyager,zheng2025skillweaver,alzubi2026evoskill}.

Despite addressing the same fundamental problem, extracting reusable knowledge from interaction experience, these communities are remarkably disconnected. A citation analysis of 1{,}136 references across 22 primary papers reveals a cross-community citation rate below 1\%. Memory papers cite skill work at 0.7\% (4/566); skill papers cite memory work at 1.2\% (7/570). Neither skill survey \citep{jiang2026skillsurvey,xu2026agentskills} cites any memory system, and only one memory survey \citep{yang2026graphmemory} cites any skill work (only Voyager). This separation reflects a conceptual gap that limits the design of scalable agent systems.

\textbf{Scope.} We study knowledge extracted at the \emph{scaffold level}: runtime systems outside model weights. Training-time methods (RLHF~\citep{ouyang2022training}, Constitutional AI) are complementary but out of scope.

\textbf{Our key observation} is that memory extraction and skill discovery are instantiations of the same operation: \emph{compressing interaction experience into reusable knowledge} at different granularities. A \textbf{memory} system extracts structured event records (${\sim}10\times$ compression); a \textbf{skill} system extracts reusable behavioral patterns (${\sim}100\times$); a \textbf{rule} system extracts abstract decision principles (${\sim}1{,}000\times$+). These are not three separate problems; they are three points on a single \textbf{experience compression spectrum}, where higher compression directly translates to reduced context consumption, faster retrieval, and lower compute overhead per decision.

This observation has cognitive science analogs. Complementary Learning Systems (CLS) theory \citep{mcclelland1995complementary} describes how the hippocampus rapidly encodes episodic memories that are gradually consolidated into neocortical knowledge, a biological compression spectrum that implies agent systems should perform upward compression during idle time. ACT-R's declarative--procedural distinction \citep{anderson1983act} maps onto our $L_1$/$L_2$ boundary, and Fitts \& Posner's \citeyearpar{fitts1967human} skill acquisition theory shows that knowledge flows \emph{bidirectionally} (explicit rules compile into automatic procedures through practice), a property no current system supports.

Practitioners already perform full-spectrum compression manually: hundreds of thousands of developers maintain \texttt{CLAUDE.md} and \texttt{.cursorrules} files that distill deployment experience into reusable rules ($L_0 \to L_3$). Yet no system automates this process; each operates at a single, predetermined compression level.

\textbf{Contributions.} We (1)~formalize the Experience Compression Spectrum, unifying agent memory, skills, and rules on a single compression axis (\cref{sec:spectrum}); (2)~map 20+ systems onto this spectrum, revealing that none supports adaptive cross-level compression, the ``missing diagonal'' (\cref{sec:spectrum}); (3)~quantify the community disconnect ($<$1\% cross-citation) and expose four structural insights invisible from within either community (\cref{sec:insights}); and (4)~articulate open problems and design principles for scalable, full-spectrum agent learning (\cref{sec:open}).

\section{The Experience Compression Spectrum}
\label{sec:spectrum}

We formalize a four-level spectrum characterizing how interaction experience is progressively compressed into increasingly abstract and reusable knowledge.

\subsection{Formal Framework}

\begin{definition}[Interaction Trace]
An interaction trace $\mathcal{T} = \{(s_t, a_t, o_t, f_t)\}_{t=1}^{N}$ is a sequence of states~$s_t$, actions~$a_t$, observations~$o_t$, and feedback signals~$f_t$ collected during agent execution.
\end{definition}

\begin{definition}[Experience Compression Function]
An experience compression function $C_L: \mathcal{T} \rightarrow \mathcal{K}_L$ maps traces to knowledge artifacts at compression level $L \in \{0,1,2,3\}$.
\end{definition}

The four levels are:

\paragraph{Level 0: Raw Trace.} The uncompressed interaction record. Format: complete logs, execution trajectories. Compression ratio: 1:1. Reusability: minimal, entirely context-bound.

\paragraph{Level 1: Episodic Memory.} Structured extraction of \emph{what happened}, preserving key contextual details while discarding redundant interaction mechanics. Format: key-value pairs, timestamped event summaries (e.g., ``[2026-03-15] User requested Q3 revenue analysis via SQL. Preferred tabular format.''). Compression ratio: ${\sim}5$--$20\times$. Reusability: low to moderate, tied to specific episodes.

\paragraph{Level 2: Procedural Skill.} Extraction of \emph{how to act} in a class of situations, abstracting across instances into reusable behavioral patterns. Format: structured routines, code snippets, workflow templates (e.g., ``\textsc{Data\_Analysis}: (1)~Confirm source, (2)~Select tool, (3)~Present in preferred format, (4)~Verify.''). Compression ratio: ${\sim}50$--$500\times$. Reusability: high, transferable across similar situations.

\paragraph{Level 3: Declarative Rule.} Extraction of \emph{what principles govern decisions}: domain-invariant knowledge. Format: natural language principles, constraints, policies (e.g., ``Always verify computed results against source data before presenting.''). Compression ratio: ${\sim}1000\times{+}$. Reusability: highest, domain-general but may lack actionable specificity.

\subsection{Properties of the Spectrum}

The four levels exhibit systematic trade-offs along three dimensions, each with direct efficiency implications.

\textbf{Generalizability vs.\ specificity.} As compression increases ($L_0 \to L_3$), knowledge becomes more broadly applicable but less context-specific.

\textbf{Compression ratio vs.\ information retention.} Higher levels discard more contextual detail via \emph{semantic abstraction}: identifying which patterns generalize and which are incidental. As concrete examples: Mem0 \citep{chhikara2025mem0} compresses multi-session conversation history (${\sim}26{,}000$ tokens) into retrieved memory entries (${\sim}1{,}800$ tokens), roughly $15\times$ at $L_1$. Trace2Skill \citep{ni2026trace2skill} distills traces from 200 tasks via 128 parallel sub-agents into a compact skill directory, roughly $100$--$500\times$ at $L_2$.

\textbf{Acquisition cost vs.\ maintenance cost.} $L_1$ memories are cheap to acquire (single trace) but expensive to maintain at scale: an $L_1$-only system accumulating thousands of entries per day will exhaust any practical retrieval budget within weeks. $L_3$ rules require many traces to induce but form a compact, low-maintenance set. This trade-off makes upward compression not merely desirable but \emph{necessary} for agents that operate at scale over extended deployments. To quantify: an $L_1$-only agent storing 1{,}000 episodes at ${\sim}500$ tokens each maintains a ${\sim}500$K-token knowledge store that must be indexed and searched at every decision. Compressing to $L_2$ skills reduces this to ${\sim}5$K tokens; to $L_3$ rules, ${\sim}500$ tokens: a $100$--$1{,}000\times$ reduction in storage and retrieval overhead that compounds across thousands of daily decisions.

Critically, these are not sequential pipeline stages. A system could compress directly from $L_0 \to L_3$, or maintain knowledge at multiple levels simultaneously. The spectrum describes the \emph{output space}, not a fixed processing order.

\subsection{Mapping Existing Systems}
\label{sec:mapping}

We select systems that (a)~learn from interaction traces (excluding pre-specified rules), (b)~produce persistent knowledge artifacts, and (c)~have been published since 2023. We position 20+ systems on the spectrum (\cref{fig:spectrum}, \cref{tab:mapping}), revealing that each operates at a fixed, predetermined compression level.\footnote{Our selection is not exhaustive; additional $L_1$ systems such as LightMem~\citep{chen2025lightmem} further corroborate the clustering pattern.}

\paragraph{Level 1 (Episodic Memory).}
Ten systems cluster here (\cref{tab:mapping}), spanning diverse mechanisms (LLM-driven extraction \citep{chhikara2025mem0}, agentic indexing \citep{xu2025amem}, hierarchical storage \citep{kang2025memoryos}, RL-optimized memory operations \citep{zhang2026memskill,yan2025memoryr1,wang2025memalpha,li2026mempo}, meta-learned memory architectures \citep{xiong2026alma}, multi-agent coordination \citep{lin2026memma}, and governance with temporal decay \citep{lam2026ssgm}) but converging on the same output: structured episodic records.

\paragraph{Level 2 (Procedural Skill).}
Eight systems cluster here. Voyager \citep{wang2023voyager} pioneered the paradigm; CASCADE \citep{huang2025cascade} chains cumulative skill creation with autonomous evolution. An emerging consensus reinforces our framework: \emph{distilling traces into abstract skills consistently outperforms storing them in retrieval memory banks} (\cref{tab:evidence}). SkillRL \citep{xia2026skillrl} reports $+68.5$pp over $L_1$ trajectory retrieval on ALFWorld; Trace2Skill \citep{ni2026trace2skill} outperforms human-written skills by $+21.5$pp on SpreadsheetBench. SkillsBench~\citep{li2026skillsbench} further finds that curated skills help ($+16.2$pp) while LLM-self-generated skills provide no benefit ($+0.0$pp): compression level alone is insufficient; the \emph{fidelity} of the compression process determines whether the artifact is useful or merely compact noise.

\paragraph{Cross-level ($L_1 \leftrightarrow L_2$).}
ExpeL \citep{zhao2024expel} and AutoAgent \citep{wang2026autoagent} operate at two levels simultaneously, but both use \emph{predetermined} levels without adaptive selection: two-speed systems, not continuous spectra.

\paragraph{Level 3 (Declarative Rule).}
Notably sparse. No surveyed system automates rule extraction from agent experience. Constitutional AI \citep{bai2022constitutional} uses \emph{pre-specified} rules; reward design methods (PBRS~\citep{ng1999pbrs}, process rewards~\citep{lightman2023prm}, rule-based rewards~\citep{shao2024deepseekmath}) encode $L_3$-type knowledge but are human-designed, not learned. Weight-level rules (via RLHF) are static after training, opaque to inspection, and cannot be updated without retraining. Scaffold-level rules would be inspectable, editable, and deployable without gradient updates.

The barriers to automated $L_3$ extraction are technical: (i)~distinguishing causal regularities from incidental correlations is harder than episode extraction; (ii)~rules without $L_2$ grounding risk being too abstract; (iii)~no methodology exists for evaluating rule quality (infinite regress of meta-evaluation); and (iv)~LLM-as-Judge gives a false sense that evaluation is solved. However, recent empirical work provides initial $L_3$ evidence: RuleShaping \citep{zhang2026agentrules} studies 25{,}000+ natural-language rules for coding agents and finds that negative constraints (guardrails) improve performance by 7--14pp while positive directives hurt, suggesting that $L_3$ compression must preserve \emph{shaping} signals rather than prescriptive instructions. SEVerA \citep{banerjee2026severa} adds formal verification to self-evolving agents, orthogonal to compression level but critical for correctness. Closing this loop at $L_3$, verbal reinforcement learning with explicit insight governance~\citep{cui2026feedbackloop} pairs rule extraction with a persistent evidence log and a curation policy, showing that the same accumulated experience either degrades performance below a zero-shot baseline or improves it depending on whether the governance loop is present.

\begin{table}[t]
\caption{Cross-level performance evidence aggregated from published benchmarks. Each row compares a higher-compression representation against a lower-compression baseline within the same study. Results span different benchmarks and cannot be directly compared across rows, but the direction is consistent: higher compression yields better downstream performance.}
\label{tab:evidence}
\centering
\footnotesize
\begin{tabular}{@{}lllr@{}}
\toprule
\textbf{Study} & \textbf{Comparison} & \textbf{Bench.} & \textbf{$\Delta$} \\
\midrule
SkillRL & $L_2$ vs.\ $L_1$ retrieval & ALFWorld & $+68.5$pp \\
Trace2Skill & $L_2$ vs.\ human skill & SpreadSh. & $+21.5$pp \\
Trace2Skill & $L_2$ vs.\ no skill & SpreadSh. & $+42.1$pp \\
SkillsBench & curated $L_2$ vs.\ none & multi-task & $+16.2$pp \\
SkillsBench & self-gen $L_2$ vs.\ none & multi-task & $+0.0$pp \\
EvoSkill & $L_2$ vs.\ no skill & BrowseC. & $+5.3$\% \\
RuleShaping & $L_3$ constr.\ vs.\ 0-shot & SWE-b. & $+7$--$14$pp \\
\bottomrule
\end{tabular}
\end{table}

\begin{table}[t]
\caption{Representative systems mapped onto the Experience Compression Spectrum. \textbf{Level} indicates primary output granularity. \textbf{Mechanism} indicates the core compression method. \textbf{Lifecycle} indicates explicit support for knowledge maintenance (versioning, conflict detection, deprecation). Notably, no system supports adaptive level selection; all operate at a predetermined compression level.}
\label{tab:mapping}
\centering
\small
\begin{tabular}{@{}llcc@{}}
\toprule
\textbf{System} & \textbf{Level} & \textbf{Mechanism} & \textbf{Lifecycle} \\
\midrule
MemSkill & $L_1$ & RL & partial \\
Mem0 & $L_1$ & LLM & partial \\
A-MEM / MemoryOS & $L_1$ & LLM & \texttimes \\
Memory-R1 / Mem-$\alpha$ & $L_1$ & RL & partial \\
MemPO & $L_1$ & RL & \texttimes \\
ALMA & $L_1$ & meta-learn & \texttimes \\
MemMA & $L_1$ & multi-agent & partial \\
SSGM & $L_1$ & governance & partial \\
\midrule
Voyager & $L_2$ & LLM & \texttimes \\
SkillWeaver & $L_2$ & LLM & \texttimes \\
EvoSkill & $L_2$ & search & partial \\
CASCADE & $L_2$ & LLM & \texttimes \\
AutoSkill & $L_2$ & LLM & \checkmark \\
Trace2Skill & $L_2$ & LLM & \texttimes \\
SkillRL & $L_2$ & RL & \texttimes \\
EvolveR & $L_2$ & RL & partial \\
\midrule
ExpeL & $L_1$--$L_2$ & LLM & partial \\
AutoAgent & $L_1$--$L_2$ & LLM & partial \\
\midrule
SEVerA & orthogonal & verify & \texttimes \\
\bottomrule
\end{tabular}
\end{table}

\subsection{The Missing Diagonal}
\label{sec:diagonal}

The mapping reveals a striking pattern: systems cluster at Level~1 or Level~2, with minimal cross-level work and virtually no Level~3 (\cref{fig:spectrum}). We call this the \textbf{missing diagonal}: the absence of systems that can:
\begin{enumerate}
  \item \textbf{Adaptively select} the appropriate compression level $L$ for a given trace $\mathcal{T}$.
  \item \textbf{Promote knowledge upward} ($C_{L \to L'}$ for $L' > L$) when sufficient evidence accumulates (many memories $\to$ one skill $\to$ one rule).
  \item \textbf{Demote knowledge downward} ($C_{L \to L'}$ for $L' < L$) when a rule proves too abstract for a specific context.
\end{enumerate}

This gap reflects a \emph{problem framing} limitation (each community defines its output format a priori), not missing engineering effort. Adaptive level selection is a \textbf{meta-learning} problem: learning \emph{what kind of knowledge} to extract, not just \emph{what knowledge} to extract. It is also a \emph{scalability bottleneck}: an $L_1$-only system accumulating episodic memories faces linear growth in storage and retrieval cost, eventually degrading performance as irrelevant entries dilute useful knowledge. Upward compression ($L_1 \to L_2 \to L_3$) is the natural solution: consolidate recurring patterns into compact skills or rules that scale sub-linearly with experience volume.

\section{What the Spectrum Reveals}
\label{sec:insights}

Viewing existing work through the compression lens exposes four structural insights invisible from within either community alone.

\subsection{Specialization Alone Is Insufficient}

One might argue that the memory--skill separation reflects healthy specialization: memory systems serve personalization, skill systems serve capability. But the spectrum reveals that both communities independently solve the same sub-problems (retrieval over growing knowledge stores, conflict detection between contradictory entries, staleness recognition, and evaluation of downstream utility) without sharing solutions. Moreover, real deployments require \emph{both} personalization and capability: an agent that abstracts reusable workflows but forgets user preferences is incomplete; one that remembers context but cannot consolidate recurring patterns will overwhelm its retrieval budget (\cref{sec:diagonal}). The spectrum makes these shared challenges visible, suggesting that a unified architecture, with level-specific compressors sharing common retrieval, conflict-resolution, and lifecycle infrastructure, would avoid redundant engineering across communities.

\subsection{Evaluation Methods Are Level-Coupled}

$L_1$ systems evaluate via QA metrics (F1, exact match), $L_2$ via task success rate, and $L_3$ has no established methodology. This coupling means systems are optimized for their level's metric, which may not reflect true downstream utility: MemSkill achieves strong QA~F1 but its impact on multi-step task completion is unclear; EvoSkill's task success metric cannot capture user-specific context. A unified evaluation framework should assess knowledge artifacts by their \textbf{downstream impact on agent performance across time}, regardless of compression level: an efficiency metric that captures value per token of stored knowledge.

\subsection{Transferability Increases with Compression Level}

Empirical evidence supports a monotonic relationship: $L_1$ memories transfer across base models (MemSkill: LLaMA $\to$ Qwen); $L_2$ skills transfer across tasks (EvoSkill: SealQA $\to$ BrowseComp, $+5.3\%$) and model sizes (Trace2Skill: 35B $\to$ 122B, $+57.7$ pp)~\citep{ni2026trace2skill}. This relationship likely follows a concave curve, with the \textbf{sweet spot} at $L_2$ balancing transferability with specificity. A controlled experiment holding source experience constant while varying only compression level has not been conducted, but existing evidence is consistent (\cref{tab:evidence}): SkillRL reports $+68.5$pp higher success with $L_2$ skills than $L_1$ trajectory retrieval on ALFWorld~\citep{xia2026skillrl}.

\subsection{Lifecycle Management Remains an Afterthought}

Existing systems focus on knowledge \emph{acquisition} while treating \emph{maintenance} as secondary. Some include partial mechanisms (Mem0's LLM-driven operations, Memory-R1's RL-learned updates, ExpeL's importance counting, AutoSkill's \citep{yang2026autoskill} version management) but these remain isolated. \textbf{No system addresses the full lifecycle}: conflict detection across knowledge types, staleness recognition, principled deprecation, and cross-level consistency. Merging skills does not always help (including success-trace signals can cause $\pm$21pp performance swings~\citep{ni2026trace2skill}), and agent behavior can degrade over extended operation even when correctness is maintained. A recent diagnosis of skill-library drift~\citep{zhang2026librarydrift} shows that without outcome-driven retirement and a bounded active-cap, self-evolving libraries degrade below their no-skill baseline, isolating lifecycle management as the failure's root cause rather than skill authoring. The software engineering community's version control and deprecation practices offer a concrete opportunity for transfer to knowledge artifacts.

\paragraph{Testable predictions.}
The spectrum generates predictions that distinguish it from pure taxonomy: (i)~$L_2$ compression should outperform $L_1$ retrieval on cross-domain transfer when holding source experience constant; (ii)~a multi-level knowledge store ($L_1$+$L_2$) should outperform either level alone, with the gap widening as deployment length increases; (iii)~the transferability--specificity curve should be concave, with $L_2$ at the sweet spot; and (iv)~$L_3$ rules should help most when framed as constraints rather than directives, consistent with early evidence from RuleShaping~\citep{zhang2026agentrules}. Validating or falsifying these predictions would establish whether the spectrum is a useful design tool or merely a descriptive framework.

\section{Open Problems and Design Principles}
\label{sec:open}

The unified perspective exposes concrete research problems, three of which we consider most pressing for building scalable agent learning systems.

\paragraph{Problem 1: Adaptive level selection.}
Design a meta-controller that, given a new trace $\mathcal{T}$, determines the optimal compression level(s) $L^*$. Recent agentic RL surveys~\citep{zhang2026agenticrlsurvey} find that RL primarily amplifies existing capabilities; genuinely new strategies emerge only under stringent conditions, suggesting that the \emph{level} of compression may matter more than the algorithm. This requires a \emph{value-of-information} framework: how much future utility does $\mathcal{K}_{L}$ provide relative to the cost of computing and maintaining it? A key design tension is whether to jointly optimize the meta-controller and level-specific compressors, or to fix the meta-controller and optimize only the compressors. Analysis of iterative generative optimization~\citep{nie2026igo} shows that design decisions (starting artifacts, credit horizons, experience batching) often dominate algorithm choice, suggesting that fixing the meta-strategy while optimizing components may be more practical.

\paragraph{Problem 2: Cross-level consistency.}
When knowledge exists at multiple levels simultaneously (e.g., an $L_1$ memory and an $L_3$ rule about the same behavior), how should the system detect and resolve inconsistencies? The analogy to database normalization is suggestive but incomplete: knowledge artifacts are semantic, not relational.

\paragraph{Problem 3: Principled lifecycle management.}
Borrow from software engineering: version control for knowledge artifacts, dependency tracking between levels, deprecation protocols, and conflict resolution. Recent workflow optimization surveys~\citep{yue2026workflowsurvey} advocate a ``minimum plasticity'' principle; an analogous principle for knowledge management, promoting only when evidence warrants, would reduce maintenance burden while preserving adaptability.

\medskip
\noindent Additional open directions include reward-free compression (extending self-training methods like STaR~\citep{zelikman2022star} and ReST~\citep{gulcehre2023rest} to guide multi-level compression without clean reward signals), cross-domain transfer protocols (can $L_2$ skills from one deployment warm-start another?), and extension to multimodal traces (visual observations and cross-modal interactions amplify the need for efficient compression).

\medskip
We propose three design principles for systems that address these problems:
\begin{enumerate}
  \item \textbf{Level-agnostic compression core.} The compression engine should be parameterized by output level, not hard-coded. DSPy~\citep{khattab2024dspy} demonstrates that declarative specifications can be compiled into optimized pipelines at a predetermined level; a level-agnostic core would generalize this.
  \item \textbf{Bidirectional promotion/demotion.} Knowledge should flow upward (memory $\to$ skill $\to$ rule when patterns accumulate) and downward (rule $\to$ skill $\to$ memory when context demands specificity).
  \item \textbf{Continuous lifecycle governance.} Every knowledge artifact should carry metadata (provenance, confidence, usage frequency, last validation time) enabling principled maintenance.
\end{enumerate}

The spectrum also suggests \emph{idle-time consolidation}: upward compression ($L_1 \to L_2 \to L_3$) should occur during low-activity periods, analogous to hippocampal-neocortical consolidation during sleep~\citep{mcclelland1995complementary}. No current system implements this; all compress synchronously at ingestion time.

\noindent Concretely, a ``diagonal'' system might comprise three components. A \emph{meta-controller} routes each trace to one or more level-specific compressors based on novelty and frequency: a first-seen pattern stays at $L_1$; once $k$ similar $L_1$ entries accumulate, the promotion engine consolidates them into an $L_2$ skill; after cross-domain recurrence, the engine generalizes to an $L_3$ rule. A \emph{promotion/demotion engine} handles these transitions: newly promoted $L_2$ skills are validated against held-out tasks before replacing their source $L_1$ entries; demotion triggers when a skill is repeatedly retrieved but unused, or when an abstract rule fails in a specific context, reverting to fresh $L_1$ evidence collection. A \emph{lifecycle manager} tracks provenance, usage frequency, and last validation time for every artifact, enabling principled deprecation.

As a concrete example: a customer-support agent encounters a timeout on \texttt{/api/export} and stores an $L_1$ memory. After five similar episodes, promotion produces an $L_2$ skill (\textsc{Handle\_Export\_Timeout}: check batch size, reduce if ${>}1000$ rows, retry). After dozens of instances across endpoints, generalization yields an $L_3$ rule (``Timeouts on data-intensive endpoints typically stem from oversized batches''). If the rule fails in a novel context, demotion drops back to $L_1$, restarting evidence collection. Existing components (MemSkill's RL controller for routing, EvoSkill's Pareto selection for promotion, AutoSkill's versioning~\citep{yang2026autoskill} for lifecycle) could serve as building blocks for this architecture.

\paragraph{Limitations.}
Our framework focuses on scaffold-level knowledge; training-time methods (RLHF, Constitutional AI) are complementary but distinct. The framework is conceptual rather than empirically validated: the spectrum's utility as a design tool awaits experimental confirmation. The four discrete levels are a simplifying abstraction; in practice, compression may be continuous (EvolveR's~\citep{wu2025evolver} strategies straddle $L_2$/$L_3$). Our survey is a snapshot of Q1~2026; the field moves rapidly. While we focus on text-based agents, the spectrum naturally extends to multimodal experience: visual observations and cross-modal interactions amplify the need for efficient compression at every level.

\section{Conclusion}

We have proposed the Experience Compression Spectrum, revealing agent memory, skill discovery, and rule learning as points on a single compression axis. Mapping 20+ systems onto this spectrum exposed the \emph{missing diagonal} (the absence of adaptive cross-level compression) alongside the insufficiency of specialization alone, level-coupled evaluation, the transferability--specificity trade-off, and neglected lifecycle management. With 10+ papers in Q1~2026 alone and production systems increasingly managing persistent experiential knowledge~\citep{chan2026composer2}, this unification is timely: without a shared framework, two communities risk solving the same problem twice. The most promising direction is building systems that adaptively select compression granularity to match the value and generality of each experience, enabling agents that scale efficiently across long deployments.

\section*{Impact Statement}
This paper presents work whose goal is to advance the field of Machine Learning. There are many potential societal consequences of our work, none which we feel must be specifically highlighted here.

\section*{LLM/Agent Usage Disclosure}
LLMs were used to polish prose and improve clarity of writing. All intellectual contributions (framework design, system analysis, citation analysis, and conclusions) are the authors' own.


\begin{thebibliography}{46}
\providecommand{\natexlab}[1]{#1}
\providecommand{\url}[1]{\texttt{#1}}
\expandafter\ifx\csname urlstyle\endcsname\relax
  \providecommand{\doi}[1]{doi: #1}\else
  \providecommand{\doi}{doi: \begingroup \urlstyle{rm}\Url}\fi

\bibitem[Alzubi et~al.(2026)Alzubi, Provenzano, Bingham, Chen, and
  Vu]{alzubi2026evoskill}
Alzubi, S., Provenzano, N., Bingham, J., Chen, W., and Vu, T.
\newblock {EvoSkill}: Automated skill discovery for multi-agent systems.
\newblock \emph{arXiv preprint arXiv:2603.02766}, 2026.

\bibitem[Anderson(1983)]{anderson1983act}
Anderson, J.~R.
\newblock \emph{The Architecture of Cognition}.
\newblock Harvard University Press, 1983.

\bibitem[Bai et~al.(2022)Bai, Kadavath, Kundu, Askell, Kernion, Jones, Chen,
  Goldie, Mirhoseini, McKinnon, et~al.]{bai2022constitutional}
Bai, Y., Kadavath, S., Kundu, S., Askell, A., Kernion, J., Jones, A., Chen, A.,
  Goldie, A., Mirhoseini, A., McKinnon, C., et~al.
\newblock Constitutional {AI}: Harmlessness from {AI} feedback.
\newblock \emph{arXiv preprint arXiv:2212.08073}, 2022.

\bibitem[Banerjee et~al.(2026)Banerjee, Xu, and Singh]{banerjee2026severa}
Banerjee, D., Xu, C., and Singh, G.
\newblock {SEVerA}: Verified synthesis of self-evolving agents.
\newblock \emph{arXiv preprint arXiv:2603.25111}, 2026.

\bibitem[Chan et~al.(2026)Chan, Shalaby, Wettig, Sanger, Zhai, Ajay, Nair,
  Snell, Lu, Shen, Jia, Cassano, Liu, Chen, et~al.]{chan2026composer2}
Chan, A., Shalaby, A., Wettig, A., Sanger, A., Zhai, A., Ajay, A., Nair, A.,
  Snell, C., Lu, C., Shen, C., Jia, E., Cassano, F., Liu, H., Chen, H., et~al.
\newblock Composer 2 technical report.
\newblock \emph{arXiv preprint arXiv:2603.24477}, 2026.

\bibitem[Chhikara et~al.(2025)Chhikara, Khant, Aryan, Singh, and
  Yadav]{chhikara2025mem0}
Chhikara, P., Khant, D., Aryan, S., Singh, T., and Yadav, D.
\newblock {Mem0}: Building production-ready {AI} agents with scalable long-term
  memory.
\newblock \emph{arXiv preprint arXiv:2504.19413}, 2025.

\bibitem[Cui et~al.(2026)Cui, Zhang, Zhang, Shao, Shi, Wang, and
  He]{cui2026feedbackloop}
Cui, Y., Zhang, X., Zhang, Y., Shao, L., Shi, X., Wang, G., and He, P.
\newblock Closing the feedback loop: From experience extraction to insight
  governance in verbal reinforcement learning.
\newblock \emph{arXiv preprint arXiv:2606.17591}, 2026.

\bibitem[Fang et~al.(2025)Fang, Deng, Xu, Jiang, Tang, Xu, Deng, Yao, Wang,
  Qiao, Chen, and Zhang]{chen2025lightmem}
Fang, J., Deng, X., Xu, H., Jiang, Z., Tang, Y., Xu, Z., Deng, S., Yao, Y.,
  Wang, M., Qiao, S., Chen, H., and Zhang, N.
\newblock {LightMem}: Lightweight and efficient memory-augmented generation.
\newblock \emph{arXiv preprint arXiv:2510.18866}, 2025.

\bibitem[Fitts \& Posner(1967)Fitts and Posner]{fitts1967human}
Fitts, P.~M. and Posner, M.~I.
\newblock \emph{Human Performance}.
\newblock Brooks/Cole, 1967.

\bibitem[Gulcehre et~al.(2023)Gulcehre, Paine, Srinivasan, Konyushkova, Weerts,
  Sharma, Siddhant, Ahern, Wang, Gu, et~al.]{gulcehre2023rest}
Gulcehre, C., Paine, T.~L., Srinivasan, S., Konyushkova, K., Weerts, L.,
  Sharma, A., Siddhant, A., Ahern, A., Wang, M., Gu, C., et~al.
\newblock Reinforced self-training ({ReST}) for language modeling.
\newblock \emph{arXiv preprint arXiv:2308.08998}, 2023.

\bibitem[Hu et~al.(2025)Hu, Liu, Yue, Zhang, Liu, Zhu, Lin, Guo, Dou, Xi,
  et~al.]{hu2025memorysurvey}
Hu, Y., Liu, S., Yue, Y., Zhang, G., Liu, B., Zhu, F., Lin, J., Guo, H., Dou,
  S., Xi, Z., et~al.
\newblock Memory in the age of {AI} agents.
\newblock \emph{arXiv preprint arXiv:2512.13564}, 2025.

\bibitem[Huang et~al.(2025)Huang, Chen, Fei, Li, Schwaller, and
  Ceder]{huang2025cascade}
Huang, X., Chen, J., Fei, Y., Li, Z., Schwaller, P., and Ceder, G.
\newblock {CASCADE}: Cumulative agentic skill creation through autonomous
  development and evolution.
\newblock \emph{arXiv preprint arXiv:2512.23880}, 2025.

\bibitem[Jiang et~al.(2026)Jiang, Li, Deng, Ma, Wang, Wang, and
  Yu]{jiang2026skillsurvey}
Jiang, Y., Li, D., Deng, H., Ma, B., Wang, X., Wang, Q., and Yu, G.
\newblock {SoK}: Agentic skills -- beyond tool use in {LLM} agents.
\newblock \emph{arXiv preprint arXiv:2602.20867}, 2026.

\bibitem[Kang et~al.(2025)Kang, Ji, Zhao, and Bai]{kang2025memoryos}
Kang, J., Ji, M., Zhao, Z., and Bai, T.
\newblock Memory {OS} of {AI} agent.
\newblock \emph{arXiv preprint arXiv:2506.06326}, 2025.

\bibitem[Khattab et~al.(2024)Khattab, Singhvi, Maheshwari, Zhang, Santhanam,
  Vardhamanan, Haq, Sharma, Joshi, Mober, et~al.]{khattab2024dspy}
Khattab, O., Singhvi, A., Maheshwari, P., Zhang, Z., Santhanam, K.,
  Vardhamanan, S., Haq, S., Sharma, A., Joshi, T.~T., Mober, H., et~al.
\newblock {DSPy}: Compiling declarative language model calls into
  self-improving pipelines.
\newblock \emph{arXiv preprint arXiv:2310.03714}, 2024.

\bibitem[Lam et~al.(2026)Lam, Li, Zhang, and Zhao]{lam2026ssgm}
Lam, C., Li, J., Zhang, L., and Zhao, K.
\newblock Governing evolving memory in {LLM} agents: Risks, mechanisms, and the
  stability and safety governed memory ({SSGM}) framework.
\newblock \emph{arXiv preprint arXiv:2603.11768}, 2026.

\bibitem[Li et~al.(2026{\natexlab{a}})Li, Zhang, Yu, Duan, Li, Xiang, Liao,
  Guo, Li, and Suo]{li2026mempo}
Li, R., Zhang, X., Yu, H., Duan, S., Li, X., Xiang, W., Liao, C., Guo, X., Li,
  Y., and Suo, J.
\newblock {MemPO}: Self-memory policy optimization for long-horizon agents.
\newblock \emph{arXiv preprint arXiv:2603.00680}, 2026{\natexlab{a}}.

\bibitem[Li et~al.(2026{\natexlab{b}})Li, Chen, Liu, Zheng, Chen, He, Li, You,
  Shen, Sun, et~al.]{li2026skillsbench}
Li, X., Chen, W., Liu, Y., Zheng, S., Chen, X., He, Y., Li, Y., You, B., Shen,
  H., Sun, J., et~al.
\newblock {SkillsBench}: Benchmarking how well agent skills work across diverse
  tasks.
\newblock \emph{arXiv preprint arXiv:2602.12670}, 2026{\natexlab{b}}.

\bibitem[Lightman et~al.(2023)Lightman, Kosaraju, Burda, Edwards, Baker, Lee,
  Leike, Schulman, Sutskever, and Cobbe]{lightman2023prm}
Lightman, H., Kosaraju, V., Burda, Y., Edwards, H., Baker, B., Lee, T., Leike,
  J., Schulman, J., Sutskever, I., and Cobbe, K.
\newblock Let's verify step by step.
\newblock \emph{arXiv preprint arXiv:2305.20050}, 2023.

\bibitem[Lin et~al.(2026)Lin, Zhang, Lu, Liu, Tang, He, Zhang, and
  Wang]{lin2026memma}
Lin, M., Zhang, Z., Lu, H., Liu, H., Tang, X., He, Q., Zhang, X., and Wang, S.
\newblock {MemMA}: Coordinating the memory cycle through multi-agent reasoning
  and in-situ self-evolution.
\newblock \emph{arXiv preprint arXiv:2603.18718}, 2026.

\bibitem[McClelland et~al.(1995)McClelland, McNaughton, and
  O'Reilly]{mcclelland1995complementary}
McClelland, J.~L., McNaughton, B.~L., and O'Reilly, R.~C.
\newblock Why there are complementary learning systems in the hippocampus and
  neocortex: Insights from the successes and failures of connectionist models
  of learning and memory.
\newblock \emph{Psychological Review}, 102\penalty0 (3):\penalty0 419--457,
  1995.

\bibitem[Ng et~al.(1999)Ng, Harada, and Russell]{ng1999pbrs}
Ng, A.~Y., Harada, D., and Russell, S.
\newblock Policy invariance under reward transformations: Theory and
  application to reward shaping.
\newblock In \emph{Proceedings of the 16th International Conference on Machine
  Learning}, pp.\  278--287, 1999.

\bibitem[Ni et~al.(2026)Ni, Liu, Liu, Sun, Zhou, Cheng, Wang, Zhao, Jiang, and
  Jiang]{ni2026trace2skill}
Ni, J., Liu, Y., Liu, X., Sun, Y., Zhou, M., Cheng, P., Wang, D., Zhao, E.,
  Jiang, X., and Jiang, G.
\newblock {Trace2Skill}: Distill trajectory-local lessons into transferable
  agent skills.
\newblock \emph{arXiv preprint arXiv:2603.25158}, 2026.

\bibitem[Nie et~al.(2026)Nie, Daull, Kuang, Akkiraju, Chaudhuri, Piasevoli,
  Rong, Yuan, Choudhary, Xiao, Fakoor, Swaminathan, and Cheng]{nie2026igo}
Nie, A., Daull, X., Kuang, Z., Akkiraju, A., Chaudhuri, A., Piasevoli, M.,
  Rong, R., Yuan, Y., Choudhary, P., Xiao, S., Fakoor, R., Swaminathan, A., and
  Cheng, C.-A.
\newblock Understanding the challenges in iterative generative optimization
  with {LLMs}.
\newblock \emph{arXiv preprint arXiv:2603.23994}, 2026.

\bibitem[Ouyang et~al.(2022)Ouyang, Wu, Jiang, Almeida, Wainwright, Mishkin,
  Zhang, Agarwal, Slama, Ray, et~al.]{ouyang2022training}
Ouyang, L., Wu, J., Jiang, X., Almeida, D., Wainwright, C., Mishkin, P., Zhang,
  C., Agarwal, S., Slama, K., Ray, A., et~al.
\newblock Training language models to follow instructions with human feedback.
\newblock \emph{Advances in Neural Information Processing Systems},
  35:\penalty0 27730--27744, 2022.

\bibitem[Packer et~al.(2023)Packer, Wooders, Lin, Fang, Patil, Stoica, and
  Gonzalez]{packer2023memgpt}
Packer, C., Wooders, S., Lin, K., Fang, V., Patil, S.~G., Stoica, I., and
  Gonzalez, J.~E.
\newblock {MemGPT}: Towards {LLMs} as operating systems.
\newblock In \emph{NeurIPS Workshop on Instruction Tuning and Instruction
  Following}, 2023.

\bibitem[Shao et~al.(2024)Shao, Wang, Zhu, Xu, Song, Zhang, Li, Wu, and
  Guo]{shao2024deepseekmath}
Shao, Z., Wang, P., Zhu, Q., Xu, R., Song, J., Zhang, M., Li, Y., Wu, Y., and
  Guo, D.
\newblock {DeepSeekMath}: Pushing the limits of mathematical reasoning in open
  language models.
\newblock \emph{arXiv preprint arXiv:2402.03300}, 2024.

\bibitem[Wang et~al.(2023)Wang, Xie, Jiang, Mandlekar, Xiao, Zhu, Fan, and
  Anandkumar]{wang2023voyager}
Wang, G., Xie, Y., Jiang, Y., Mandlekar, A., Xiao, C., Zhu, Y., Fan, L., and
  Anandkumar, A.
\newblock Voyager: An open-ended embodied agent with large language models.
\newblock \emph{arXiv preprint arXiv:2305.16291}, 2023.

\bibitem[Wang et~al.(2026)Wang, Liao, Wei, Tang, and Xiong]{wang2026autoagent}
Wang, X., Liao, N., Wei, S., Tang, C., and Xiong, F.
\newblock {AutoAgent}: Evolving cognition and elastic memory orchestration for
  adaptive agents.
\newblock \emph{arXiv preprint arXiv:2603.09716}, 2026.

\bibitem[Wang et~al.(2025)Wang, Takanobu, Liang, Mao, Hu, McAuley, and
  Wu]{wang2025memalpha}
Wang, Y., Takanobu, R., Liang, Z., Mao, Y., Hu, Y., McAuley, J., and Wu, X.
\newblock {Mem-$\alpha$}: Learning memory construction via reinforcement
  learning.
\newblock \emph{arXiv preprint arXiv:2509.25911}, 2025.

\bibitem[Wu et~al.(2025)Wu, Wang, Mei, Cai, Fu, Yang, Wen, Yang, Shen, Wang,
  and Shi]{wu2025evolver}
Wu, R., Wang, X., Mei, J., Cai, P., Fu, D., Yang, C., Wen, L., Yang, X., Shen,
  Y., Wang, Y., and Shi, B.
\newblock {EvolveR}: Self-evolving {LLM} agents through an experience-driven
  lifecycle.
\newblock \emph{arXiv preprint arXiv:2510.16079}, 2025.

\bibitem[Xia et~al.(2026)Xia, Chen, Wang, Liu, Zeng, Wang, Han, Zhou, Zhao,
  Chen, Zheng, Xie, and Yao]{xia2026skillrl}
Xia, P., Chen, J., Wang, H., Liu, J., Zeng, K., Wang, Y., Han, S., Zhou, Y.,
  Zhao, X., Chen, H., Zheng, Z., Xie, C., and Yao, H.
\newblock {SkillRL}: Evolving agents via recursive skill-augmented
  reinforcement learning.
\newblock \emph{arXiv preprint arXiv:2602.08234}, 2026.

\bibitem[Xiong et~al.(2026)Xiong, Hu, and Clune]{xiong2026alma}
Xiong, Y., Hu, S., and Clune, J.
\newblock Learning to continually learn via meta-learning agentic memory
  designs.
\newblock \emph{arXiv preprint arXiv:2602.07755}, 2026.

\bibitem[Xu \& Yan(2026)Xu and Yan]{xu2026agentskills}
Xu, R. and Yan, Y.
\newblock Agent skills for large language models: Architecture, acquisition,
  security, and the path forward.
\newblock \emph{arXiv preprint arXiv:2602.12430}, 2026.

\bibitem[Xu et~al.(2025)Xu, Liang, Mei, Gao, Tan, and Zhang]{xu2025amem}
Xu, W., Liang, Z., Mei, K., Gao, H., Tan, J., and Zhang, Y.
\newblock {A-MEM}: Agentic memory for {LLM} agents.
\newblock \emph{arXiv preprint arXiv:2502.12110}, 2025.

\bibitem[Yan et~al.(2025)Yan, Yang, Huang, Nie, Ding, Li, Ma, Bi, Kersting,
  Pan, Sch{\"u}tze, Tresp, and Ma]{yan2025memoryr1}
Yan, S., Yang, X., Huang, Z., Nie, E., Ding, Z., Li, Z., Ma, X., Bi, J.,
  Kersting, K., Pan, J.~Z., Sch{\"u}tze, H., Tresp, V., and Ma, Y.
\newblock {Memory-R1}: Enhancing large language model agents to manage and
  utilize memories via reinforcement learning.
\newblock \emph{arXiv preprint arXiv:2508.19828}, 2025.

\bibitem[Yang et~al.(2026{\natexlab{a}})Yang, Zhou, Xiao, Dong, Zhuang, Zhang,
  Wang, Hong, Yuan, Xiang, et~al.]{yang2026graphmemory}
Yang, C., Zhou, C., Xiao, Y., Dong, S., Zhuang, L., Zhang, Y., Wang, Z., Hong,
  Z., Yuan, Z., Xiang, Z., et~al.
\newblock Graph-based agent memory: Taxonomy, techniques, and applications.
\newblock \emph{arXiv preprint arXiv:2602.05665}, 2026{\natexlab{a}}.

\bibitem[Yang et~al.(2026{\natexlab{b}})Yang, Li, Pan, Zhan, Cai, Du, Zhou,
  Chen, Chen, Li, Zhang, and He]{yang2026autoskill}
Yang, Y., Li, J., Pan, Q., Zhan, B., Cai, Y., Du, L., Zhou, J., Chen, K., Chen,
  Q., Li, X., Zhang, B., and He, L.
\newblock {AutoSkill}: Experience-driven lifelong learning via skill
  self-evolution.
\newblock \emph{arXiv preprint arXiv:2603.01145}, 2026{\natexlab{b}}.

\bibitem[Yue et~al.(2026)Yue, Bhandari, Ko, Patel, Lin, Zhou, Gao, Chen, and
  Pan]{yue2026workflowsurvey}
Yue, L., Bhandari, K.~R., Ko, C.-Y., Patel, D., Lin, S., Zhou, N., Gao, J.,
  Chen, P.-Y., and Pan, S.
\newblock From static templates to dynamic runtime graphs: A survey of workflow
  optimization for {LLM} agents.
\newblock \emph{arXiv preprint arXiv:2603.22386}, 2026.

\bibitem[Zelikman et~al.(2022)Zelikman, Wu, Mu, and Goodman]{zelikman2022star}
Zelikman, E., Wu, Y., Mu, J., and Goodman, N.~D.
\newblock {STaR}: Bootstrapping reasoning with reasoning.
\newblock \emph{Advances in Neural Information Processing Systems},
  35:\penalty0 15476--15488, 2022.

\bibitem[Zhang et~al.(2026{\natexlab{a}})Zhang, Geng, Yu, Yin, Zhang, Tan,
  Zhou, Li, Xue, Li, et~al.]{zhang2026agenticrlsurvey}
Zhang, G., Geng, H., Yu, X., Yin, Z., Zhang, Z., Tan, Z., Zhou, H., Li, Z.,
  Xue, X., Li, Y., et~al.
\newblock The landscape of agentic reinforcement learning for {LLMs}: A survey.
\newblock \emph{Transactions on Machine Learning Research (TMLR)},
  2026{\natexlab{a}}.

\bibitem[Zhang et~al.(2026{\natexlab{b}})Zhang, Long, Bao, Feng, Zhang, Yue,
  and Wang]{zhang2026memskill}
Zhang, H., Long, Q., Bao, J., Feng, T., Zhang, W., Yue, H., and Wang, W.
\newblock {MemSkill}: Learning and evolving memory skills for self-evolving
  agents.
\newblock \emph{arXiv preprint arXiv:2602.02474}, 2026{\natexlab{b}}.

\bibitem[Zhang et~al.(2026{\natexlab{c}})Zhang, Cui, Wang, Li, Qiu, Zhu, and
  He]{zhang2026librarydrift}
Zhang, X., Cui, Y., Wang, G., Li, Z., Qiu, W., Zhu, B., and He, P.
\newblock Library drift: Diagnosing and fixing a silent failure mode in
  self-evolving {LLM} skill libraries.
\newblock \emph{arXiv preprint arXiv:2605.19576}, 2026{\natexlab{c}}.

\bibitem[Zhang et~al.(2026{\natexlab{d}})Zhang, Wang, Cui, Qiu, Li, Zhu, and
  He]{zhang2026agentrules}
Zhang, X., Wang, G., Cui, Y., Qiu, W., Li, Z., Zhu, B., and He, P.
\newblock Do agent rules shape or distort? {G}uardrails beat guidance in coding
  agents.
\newblock \emph{arXiv preprint arXiv:2604.11088}, 2026{\natexlab{d}}.

\bibitem[Zhao et~al.(2024)Zhao, Huang, Xu, Lin, Liu, and Huang]{zhao2024expel}
Zhao, A., Huang, D., Xu, Q., Lin, M., Liu, Y.-J., and Huang, G.
\newblock {ExpeL}: {LLM} agents are experiential learners.
\newblock In \emph{Proceedings of the AAAI Conference on Artificial
  Intelligence}, volume~38, pp.\  19632--19642, 2024.

\bibitem[Zheng et~al.(2025)Zheng, Fatemi, Jin, Wang, Gandhi, Song, Gu,
  Srinivasa, Liu, Neubig, and Su]{zheng2025skillweaver}
Zheng, B., Fatemi, M.~Y., Jin, X., Wang, Z.~Z., Gandhi, A., Song, Y., Gu, Y.,
  Srinivasa, J., Liu, G., Neubig, G., and Su, Y.
\newblock {SkillWeaver}: Web agents can self-improve by discovering and honing
  skills.
\newblock \emph{arXiv preprint arXiv:2504.07079}, 2025.

\end{thebibliography}
\end{document}